\documentclass[letterpaper, 10 pt, conference]{ieeeconf} 
\IEEEoverridecommandlockouts 
\usepackage{cite}
\usepackage{amsmath,amssymb,amsfonts}
\usepackage{graphicx}
\usepackage{textcomp}
\usepackage{xcolor}
\def\BibTeX{{\rm B\kern-.05em{\sc i\kern-.025em b}\kern-.08em
    T\kern-.1667em\lower.7ex\hbox{E}\kern-.125emX}}
\usepackage{graphicx}
\usepackage{subfig} 
\usepackage[export]{adjustbox}

\usepackage[switch]{lineno}
\usepackage{amssymb}
\usepackage{amsmath}
\usepackage{mathrsfs}
\usepackage{hyperref}

\usepackage{multirow}

\usepackage[linesnumbered,vlined,ruled]{algorithm2e}
\usepackage{setspace}
\SetKwProg{Fn}{Function}{}{}
\SetCommentSty{emph}
\SetArgSty{textrm}
\usepackage{cleveref}
\Crefformat{figure}{#2Fig.~#1#3}
\Crefmultiformat{figure}{Figs.~#2#1#3}{ and~#2#1#3}{, #2#1#3}{ and~#2#1#3}
\usepackage{color}
\SetAlFnt{\small}
\SetAlCapFnt{\small}
\SetAlCapNameFnt{\large}
\usepackage{algorithmic}
\algsetup{linenosize=\tiny}

\newcommand{\PREC} {\boldsymbol{\pmb\prec}}

\newcommand{\True} {{\emph{true}}}
\newcommand{\False} {{\emph{false}}}

\definecolor{junglegreen}{rgb}{0.16, 0.67, 0.53}
\definecolor{purple}{rgb}{0.53,0.16,0.88}

\newcommand{\revise}[1]{{#1}}

\newcommand\blfootnote[1]{%
  \begingroup
  \renewcommand\thefootnote{}\footnote{#1}%
  \addtocounter{footnote}{-1}%
  \endgroup
}

\begin{document}


\title{
Cooperative Task and Motion Planning for Multi-Arm Assembly Systems
}

\author{Jingkai Chen$^1$, Jiaoyang Li$^{2,*}$, Yijiang Huang$^{1,*}$, Caelan Garrett$^{1,4}$, Dawei Sun$^3$, \\  Chuchu Fan$^1$, Andreas Hofmann$^{1}$,  Caitlin Mueller$^1$, Sven Koenig$^2$, Brian C. Williams$^1$
}

\maketitle

\begin{abstract}

Multi-robot assembly systems are becoming increasingly appealing in manufacturing due to their ability to automatically, flexibly, and quickly construct desired structural designs.
However, effectively planning for these systems in a manner that ensures each robot is simultaneously productive, and not idle, is challenging due to (1) the close proximity that the robots must operate in to manipulate the structure and (2) the inherent structural partial orderings on when each part can be installed.
In this paper, we present a task and motion planning framework that jointly plans safe, low-makespan plans for a team of robots to assemble complex spatial structures. 
Our framework takes a hierarchical approach that, at the high level, uses Mixed-integer Linear Programs to compute an abstract plan comprised of an allocation of robots to tasks subject to precedence constraints and, at the low level, builds on a state-of-the-art algorithm for Multi-Agent Path Finding to plan collision-free robot motions that realize this abstract plan. 
Critical to our approach is the inclusion of certain collision constraints and movement durations during high-level planning, which better informs the search for abstract plans that are likely to be both feasible and low-makespan while keeping the search tractable.
We demonstrate our planning system on several challenging assembly domains with several (sometimes heterogeneous) robots with grippers or suction plates for assembling structures with up to 23 objects involving Lego bricks, bars, plates, or irregularly shaped blocks.
\blfootnote{This work was supported by Kawasaki Heavy Industry, Ltd (KHI) under grant number 030118-00001. This article solely reflects the opinions and conclusions of its authors and not KHI or any other Kawasaki entity.}
\blfootnote{$^1$Massachusetts Institute of Technology; $^2$University of Southern California; $^3$University of Illinois Urbana-Champaign; $^4$NVIDIA; $^*$ indicates equal contributions; Email: jkchen@csail.mit.edu}
\end{abstract}

\section{Introduction}
Autonomous robots have been increasingly deployed to perform assembly tasks in factories.
However, most robotic assembly is manually programmed, usually requiring months to accommodate new production needs. 
To be more flexible and adaptable to varying robot setups and products, we consider automated planning approaches to robotic assembly.

Planning for a team of robots to assemble a structure requires both (1) assigning every manipulation task to a robot and scheduling them subject to temporal constraints as well as (2) solving for motions that perform each manipulation while avoiding collisions between the moving robots and objects.
These problems can be very challenging; even finding a feasible solution is non-trivial.
First, the space of possible solutions is large: an industrial manipulator usually has at least six degree-of-freedoms (DOFs), and the joint state space of multiple robots and objects grows exponentially in the number of entities.
Second, the assembly requirements in manufacturing typically involve a large number of different tasks ({\it e.g.,} assembling, holding, and welding) being performed over a long time horizon.
Additionally, tasks are often tightly coupled with precedence constraints and concurrency requirements given the product properties. 
Third, the robots constantly operate in close proximity to each other in order to manipulate the structure and must avoid collisions with other moving robots and objects. 
Finally, an arbitrary feasible solution is typically not sufficient; minimizing solution makespan ({\it i.e.} total assembly time) is critical for efficiency and productivity in industrial applications.



\begin{figure*}[t]
\centering
\vspace{-5pt}
\includegraphics[width=1.85\columnwidth]{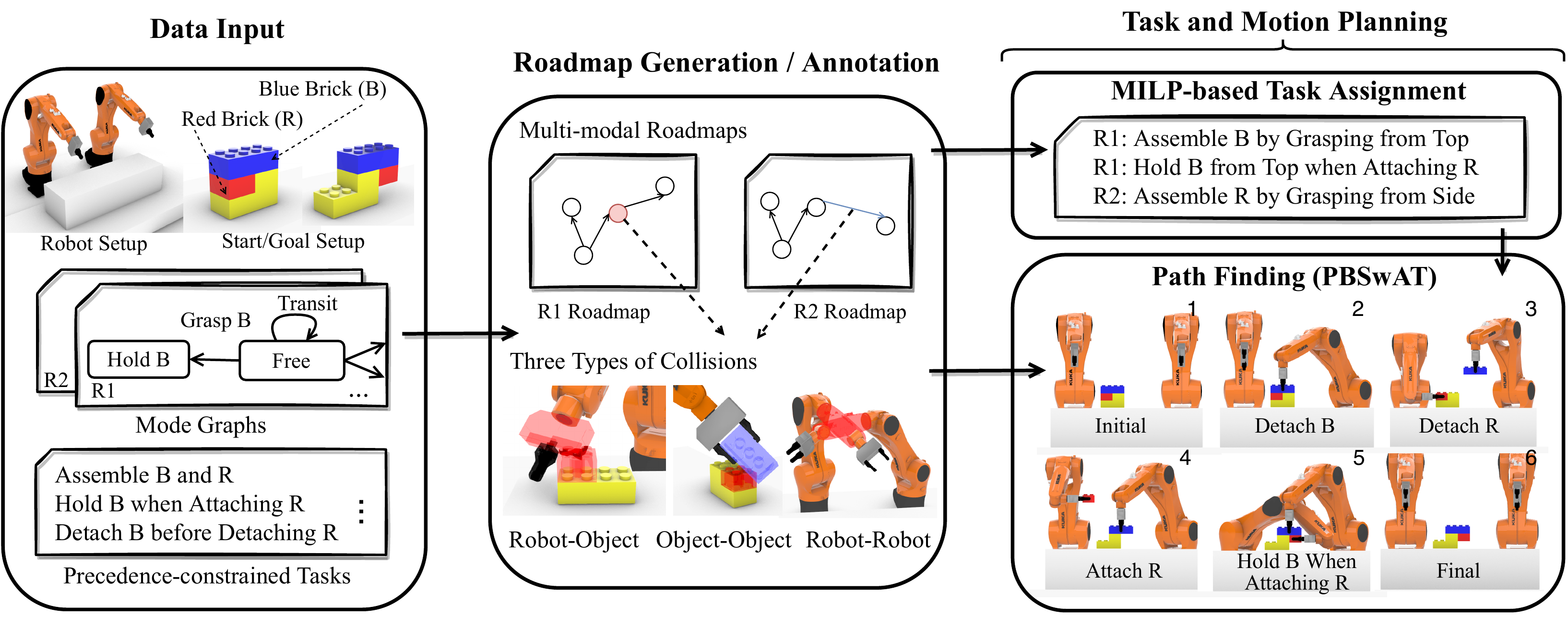}
\vspace{-4pt}
\caption{\small Phases of the proposed multi-robot task and motion planning algorithm on a two-robot Lego assembly problem.}
\label{fig:diagram}
\vspace{-16pt}
\end{figure*}

Consider the example in \Cref{fig:diagram}. Two robot arms R1 and R2 must detach and attach the bricks to move them from their start poses to their goal poses (the top left of \Cref{fig:diagram}). The precedence-constrained tasks are given in the bottom left of \Cref{fig:diagram}. In particular, after the blue brick (B) is assembled in its goal pose, one robot needs to hold B in order for the other robot to attach the red brick (R) (Steps 4 and 5 in the bottom right of \Cref{fig:diagram}).  
A planner needs to find an assignment of these tasks (the top right of \Cref{fig:diagram}) and plan paths that adhere to the assignments while avoiding three types of collisions (the bottom middle of \Cref{fig:diagram}). The bottom right of \Cref{fig:diagram} sketches six key states on a solution.

We propose a task and motion planning framework to plan safe, \revise{low-makespan} plans for multi-task multi-robot assembly problems, which addresses the above challenges. The structure of the framework is shown in \Cref{fig:diagram}. This system takes as input the robot and assembly structure setup along with descriptions of the possible robot operating modalities and a goal description that consists of a set of tasks and precedence constraints between tasks.
Our hierarchical planning system has three phases: (1) generate a multi-modal roadmap for each robot that describes its feasible movements and interactions with objects, and annotate colliding situations among different roadmaps;
(2) optimally assign assembly tasks to robots based on their roadmaps by using Mixed-Integer Linear Programming (MILP) to solve a relaxed problem that still incorporates critical collisions when robots are performing manipulations;
(3) deploy a priority-based, decoupled multi-agent search on the collision-annotated roadmaps to efficiently generate collision-free motion plans that fulfill the precedence-constrained tasks and respect the task assignments, which is biased towards plans with short makespan.
Although the framework is a strict hierarchy that does not backtrack if no collision-free motion plans are found by (3), unlike many TAMP approaches, the high-level planner (2) 
directly considers aspects such as robot reachability, critical collisions, and precedence constraints, omitting only motion decision variables and their associated collision constraints, which substantially increase the size of the mathematical program.
In practice, this not only produces feasible motion planning problems at the low level (3) but also ones that empirically admit low-makespan solutions.
Finally, we tested our planning system's ability to automatically generate \revise{low-makespan} assembly solutions for complex assembly structures with multiple robot arms on challenging simulated assembly domains involving Lego bricks, bars, plates, and irregular-shaped blocks with grippers or suction plates as end-effectors for up to 23 objects. 

\section{Related Work}

Work in Multi-Modal Motion Planning (MMMP) addresses multi-step manipulation by planning across robot configuration spaces defined by changing manipulation modes~\cite{hauser2011randomized,hauser2010multi}.
Building on MMMP, work in Task and Motion Planning (TAMP) bridges symbolic reasoning about actions that achieve discrete goals and geometric reasoning in search of collision-free robotic motions~\cite{Garrett2021}.
Most existing MMMP and TAMP methods are only able to plan for sequential systems, such as a single robot~\cite{Garrett2021} or a team of synchronized robots~\cite{shome2021synchronized}, and are unable to represent plans where manipulation can happen asynchronously, for example when one robot places an object while other robots move according to their current manipulation modes.
Existing algorithms for multi-agent TAMP~\cite{Toussaint2017Multi-boundDomains} are capable of modeling multi-arm assembly problems but have not demonstrated the ability to solve TAMP problems with the long horizons and close robot proximity that are required in multi-arm assembly.
By specializing in assembly problems, our system circumvents these challenges by using an efficient, collision-aware MILP formulation for high-level task assignment and leveraging ideas from state-of-the-art multi-agent path finding for low-level multi-modal motion planning.




Several other planning systems for multi-arm assembly have been developed 
for assembling furniture \cite{knepper2013ikeabot,dogar2019multi}, construction architecture\cite{hartmann2021long}, and LEGO bricks \cite{nagele2020legobot}.
In some systems \cite{knepper2013ikeabot,dogar2019multi}, robots are restricted to move sequentially when installing new components.
By using a decomposition approach that plans limited-horizon sub-problems in sequence, \cite{hartmann2021long} is capable of finding solutions for up to 12 robots.
However, their setting considers mobile manipulators operating in relatively open workspaces, so they are able to plan motions between identified manipulation keyframes using just a single-agent space-time motion planner.
In contrast, we focus on planning for fixed-based robots working in crowded workspaces, in which finding jointly collision-free paths is very challenging, prompting us to build on tools from multi-agent path finding. 
In a crowded assembly setting similar to ours, \cite{nagele2020legobot} plans for three fixed-base robot arms to assemble up to 32 bricks, but they are unable to plan collision-free asynchronous motion for the robots.



Multi-Agent Path Finding (MAPF) plans for a group of agents, such as vehicles or drones, to reach specified goals without colliding. 
MAPF algorithms operate on finite graphs where each agent occupies exactly one vertex and can only move to adjacent vertices at each discretized time step. 
Recent work has significantly improved the scalability of MAPF algorithms~\cite{felner2017search} and generalized grid-world planning to incorporate task assignment~\cite{brown2020optimal,honig2018conflict,zhang2022multi}. 
Conflict-based search originates from MAPF and has been successfully applied to perform multi-arm motion planning \cite{solis2021representation}.
Inspired by priority-based approaches~\cite{ma2019searching,zhang2022multi},
our system explores task priorities to generate collision-free paths. 
Compared to classical MAPF, our system is able to plan paths that achieve multiple tasks and operate on general roadmaps with continuous traversal times.



\section{Problem Formulation}

A {\em multi-arm assembly planning problem} is defined as the problem of planning trajectories
for a team of possibly heterogeneous robots $A = \{a_1,..,a_N\}$ that manipulate a set of objects $O = \{o_1,..,o_M\}$ with respect to given assembly requirements, start and finish at specified configurations, avoid collisions among robots, assembly objects, and other obstacles.
We consider holonomic robots that move subject to maximum velocity constraints. 
Each robot {\em trajectory} is a time-stamped path $[(t_0,c_0),..,(t_K,c_K)]$ of time-configuration pairs $(t_k, c_k)$. 
Velocities on the trajectory $(c_k - c_{k-1}) / (t_k - t_{k-1})$ for $t_k > t_{k-1}$
must remain within the velocity limits. We minimize the {\em makespan} $\max_{n \in [N]} t_K^n$ as our optimization criteria.

In our example in \Cref{fig:diagram}, objects B and R must be detached from the base and then attached at a new location. During the process, B needs to be held while R is being attached. To describe such assembly requirements, we use mode graphs to qualitatively describe the procedures of the robots manipulating the objects. We use precedence-constrained tasks that leverages the mode graph representation to describe the assembly requirements.
 
\begin{figure}[t!]
\centering
\includegraphics[width=0.75\columnwidth]{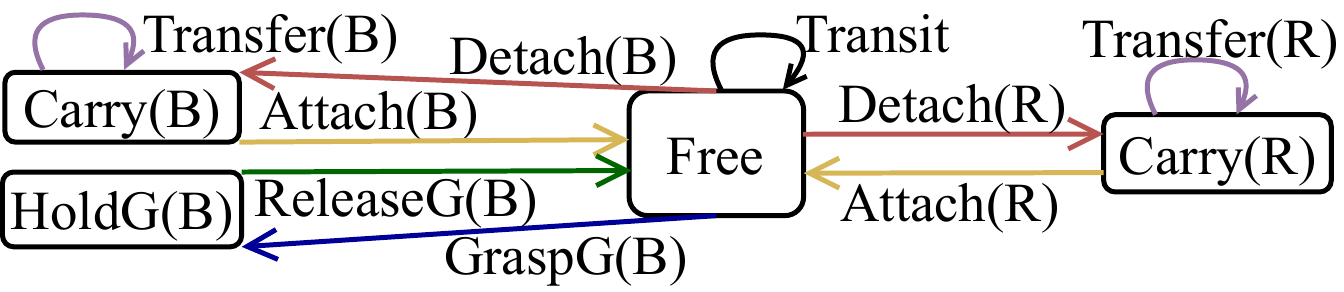}
\vspace{-3pt}
\caption{\small Mode graph.}
\label{fig:mode}
\end{figure}

A {\em mode graph} for a robot $a_n \in A$ is a directed graph $\langle \Sigma_n, \mathcal{T}_n \rangle$. Nodes $\Sigma_n$ are a set of {\em modes} that specify qualitative relations between a robot $a_n$ and the objects, each of which represents a set of robot configurations and the manipulated object states. Directed edges $\mathcal{T}_n$ between nodes are motion primitives, each of which represents a set of moving or manipulating trajectories. A trajectory along with the manipulated object states can be qualitatively described as a time-stamped, interleaving sequence of modes and primitives. \Cref{fig:mode} shows the mode graph of a robot manipulating objects B and R in our example. The modes classify states as free hand (Free), carry an object (Carry), and hold an object at its goal to be stable (HoldG). The primitives classify trajectories as move with free hand (Transit), transfer an object (Transfer), detach an object from its base or attach it at its target (Detach, Attach), and grasp or release an object at its goal (GraspG, ReleaseG).

\begin{figure}[t!]

\centering
\vspace{-10pt}
\includegraphics[width=0.85\columnwidth]{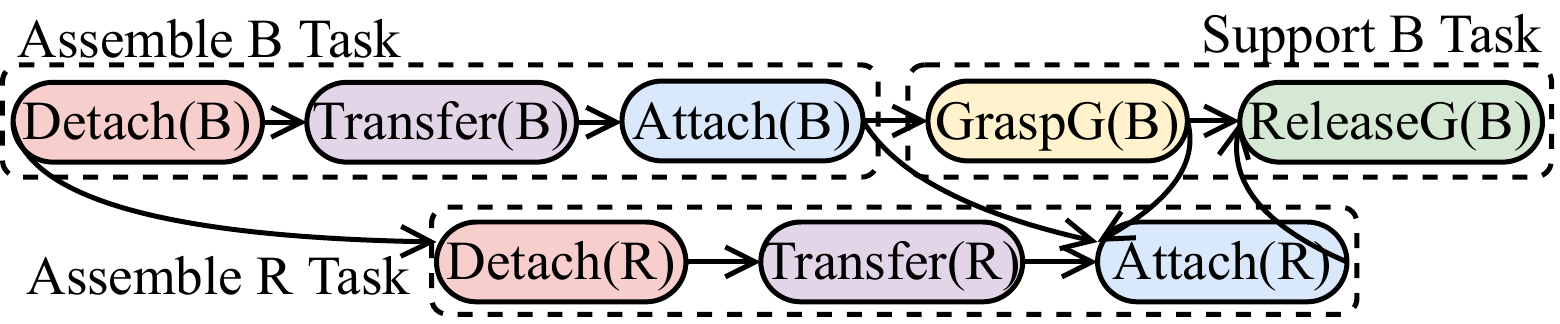}
\vspace{-3pt}
\caption{\small Precedence-constrained tasks.}
\label{fig:plan}
\vspace{-15pt}
\end{figure}

The assembly requirement is given as a set of precedence-constrained tasks $\langle \textsc{T}, \textsc{P} \rangle$ consisting of tasks $\textsc{T}$ and precedence constraints $\textsc{P}$.
Each task $\textsc{T}_p = (\tau_{p,1},..,\tau_{p,K_p}) \in \textsc{T}$ is a sequence of motion primitives that should be \revise{assigned to} and achieved by a robot. 
Each constraint $(a, b) \in P$, where $a$ and $b$ are in the form of $(\tau,\vdash)$ or $(\tau,\dashv)$, represents the start ($\vdash$) or end ($\dashv$) times of $\tau$. \revise{Precedence constraints only specify necessary partial orderings instead of a total ordering}. \Cref{fig:plan} shows an example of precedence-constrained tasks. The first two are assembly tasks with primitives [Detach(B), Transfer(B), Attach(B)] and [Detach(R), Transfer(R), Attach(R)]. The other one is a support task [GraspG(B), ReleaseG(B)]. There are three types of precedence constraints in the example: (1) the precedence constraints between subsequent modes or primitives in the same task ({\it e.g.,} Detach(B) precedes Transfer(B)); (2) Detach(B) precedes Detach(R), Attach(B) precedes Attach(R), and Attach(B) precedes GraspG(B) given the assembly structure; (3) Attach(R) precedes Grasp(B) and succeeds ReleaseG(B) given the support requirement.

In this paper, we assume: (1) each object can only be manipulated by one robot at a time, and thus lifting or handover among robots are not supported; 
(2) objects are relatively light compared to the robot payload, and objects are solid; 
(3) the assembly process is monotonic and thus objects do not need to be placed at intermediate locations for regrasping. 




\section{Roadmap Generation}
\label{section:roadmap}

\begin{figure}[t]
\centering
\includegraphics[width=0.90\columnwidth]{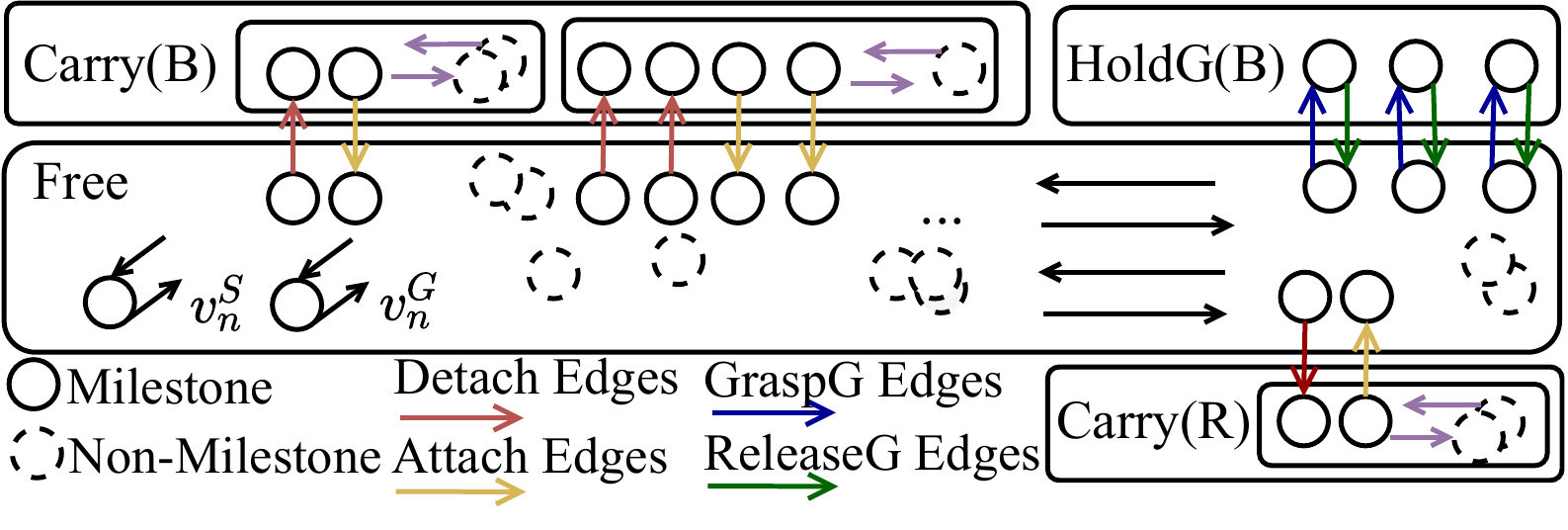}
\vspace{-5pt}
\caption{\small Multi-modal roadmap (most edges in single-mode roadmaps such as Transit-Free and Transfer-Carry are omitted).}
\label{fig:map}
\vspace{-16pt}
\end{figure}

Given a robot $a_n$ and its mode graph $(\Sigma_n, \mathcal{T}_n)$, we generate a multi-modal Probabilistic Roadmap $G_n = (\textsc{V}_n, \textsc{E}_n)$ that describes its feasible movements and interactions with the objects. Vertices $\textsc{V}_n$ are a set of configurations. Each vertex $v \in \textsc{V}_n$ is labeled with a mode $v.\sigma$ and, when relevant, \revise{a grasp pose $v.\texttt{p}$ that describes the relative transformation between the robot's end-effector and the manipulated object}. 
Let $v.\texttt{p} = \emptyset$ if $v.\sigma = \text{Free}$. 
Directed edges $\textsc{E}_n$ are a set of configuration trajectories. Each edge $e \in \textsc{E}_n$ from vertex $e.\texttt{s}$ to vertex $e.\texttt{e}$ is labelled with \revise{a traversal time $e.\texttt{w}$ of this configuration trajectory given the maximum joint velocities}, a primitive $e.\tau$ and a grasp pose $e.\texttt{p}$. 
Let $e.\texttt{p} = \emptyset$ if $e.\tau = \text{Transit}$. These vertices and edges are generated to be free of robot self-collisions and collisions with static obstacles. ~\Cref{fig:map} shows an example multi-modal roadmap.

We sample the multi-modal roadmap in a manner similar to the general MMMP sampling method in \cite{hauser2010multi}, which iteratively samples in the mode configuration spaces and their \revise{intersections}. In our mode graph, primitives are either mode-changing ({\it e.g.,} Detach, Attach, GraspG, ReleaseG) or mode-preserving ({\it e.g.,}, Transit, Transfer).
We take a task-aware sampling method to take advantage of the structure of assembly problems.
Edges and vertices are sampled as follows: 
(1) we first use manipulation skill samplers to generate a diverse set of trajectories as edges called \textit{mode-changing edges} ({\it i.e.} the non-black edges in \Cref{fig:map}) for mode-changing motion primitives, and their starts and ends are \textit{milestones} of the corresponding modes ({\it e.g.,} the solid-line circles); 
(2) then we sample \textit{mode-preserving edges} and vertices for the mode-preserving primitives and the pointed modes, which also connect the previously sampled milestones. 

These edges and vertices compose a \textit{single-mode roadmap} ({\it e.g.,} Transit-Free roadmap and Transfer-Carry roadmap) of a mode-preserving primitive and its pointed mode ({\it e.g.,} the black lines and the dashed circles). 
We have two ways to sample single-mode roadmaps: (2a) we use roadmap spanners \cite{kavraki1996probabilistic}, which are connected to the milestones; (2b) we also use RRT-Connect \cite{kuffner2000rrt} to find paths ({\it i.e.} an interleaving sequence of edges and vertices) between milestones as \textit{highways}. Vertices in (2a) and (2b) are connected together via \textit{connection edges} to enhance connectivity. Because roadmap vertices also differ in grasp poses, a single-mode roadmap can have disconnected components under different poses. For example, grasping B from the top or side results in two disjoint components in Carry(B), as in Fig.\ref{fig:map}. 

In a multi-modal roadmap, mode-changing edges (step 1) and highways (step 2b) capture the fastest paths for a robot to complete tasks while the spanned vertices and edges (step 2a) serve as alternatives when the highways are blocked by other robots during a time window. To reduce the roadmap size, we sample and add these edges differently for task assignment and path finding. In task assignment, since we only consider inter-robot collisions of mode-changing edges, we compute a multi-modal roadmap only consisting of mode-changing edges and highways. Then, when the tasks are assigned, we compute a multi-modal roadmap consisting of all the spanned vertices, spanned edges, and highways that are related to the assigned tasks, which are connected via corresponding connection edges. As each object has a unique Carry-Transfer roadmap, the number of spanned vertices and edges increases linearly in the number of objects. Thus, to further reduce the generation time, we cache the arm configurations and collision information in the Transit-Free roadmap and reuse them for spanning other single-mode roadmaps for the same robot. As a result, a large number of spanned components share the same arm configurations.

\subsection{Collision Annotation}

Although each robot and its manipulated object are guaranteed to avoid collisions with the fixed obstacles when traversing its roadmap, they also need to avoid colliding with the other robots and objects.
We adopt the idea of annotated collisions $\Pi$ in \cite{honig2018trajectory} to characterize pairwise collisions between robot and robot, robot and object, or object and object in our assembly problem. Each annotated collision $\pi \in \Pi$ is a pair of conditions, where each condition denotes an area swept by a robot or an object in the workspace, and the annotated collision implies that the two areas overlap.  
Specifically, we have two conditions types: (1) edge condition $(a_n, e)$ and vertex condition $(a_n, v)$ represent the swept area of robot $a_n$ and its manipulated object when traversing edge $e \in \textsc{E}_n$ or waiting at vertex $v \in \textsc{V}_n$ respectively; 
(2) object condition $(o_m, \bot)$ or $(o_m, \top)$ represents the area occupied by object $o_m$ being at its start ($\bot$) or goal pose ($\top$) respectively. 
With all the roadmaps, we collect the conditions for all the robots and objects. Then, we do pairwise collision checks between them to record the colliding pairs as annotated collisions. \revise{We say there is a collision when a pair of annotated conditions both hold true at the same time.} 
As a large portion of vertices and edges in roadmaps share the same arm configurations, they sweep the same area, and the collisions between them and others are only checked once. 

\section{Task Assignment}

\label{section:task}
The task assignment module generates a plan in which each robot performs a sequence of tasks such that all assembly tasks are assigned with respect to the assembly requirements and roadmap connectivity. The optimizing criteria is minimizing the plan makespan. The task assignment problem at this stage is a 
relaxation of the full problem since we ignore potential collisions when robots traverse through non-milestones. 
However, the inclusion of some collision constraints as well as non-trivial lower bounds on path-traversal durations ensures that this relaxation is representative of the full problem.
Fully collision-free paths will be generated by our multi-task multi-agent path finding algorithm by refining a set of partially ordered, unscheduled subplans extracted from our task assignment solution (Section~\ref{section:path}). This task assignment problem can be treated as an extension of Vehicle Routing Problems with Time Windows with exclusion constraints \cite{laporte1995routing} and formulated as a Mixed-Integer Linear Program (MILP) \cite{brown2020optimal}. 
\begin{figure}[t]
\centering
\subfloat[Task roadmap]{\includegraphics[height=0.28\columnwidth]{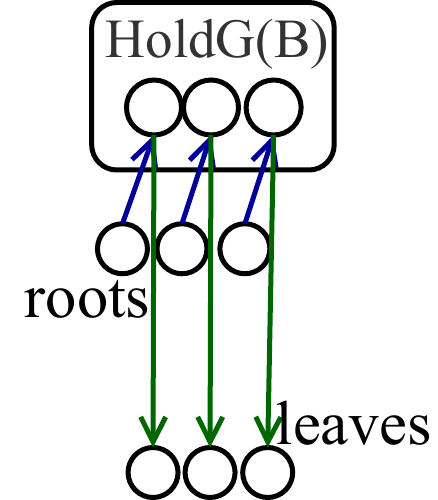}}
\,
\vspace{-5pt}
\subfloat[Plan roadmap]{\includegraphics[height=0.33\columnwidth]{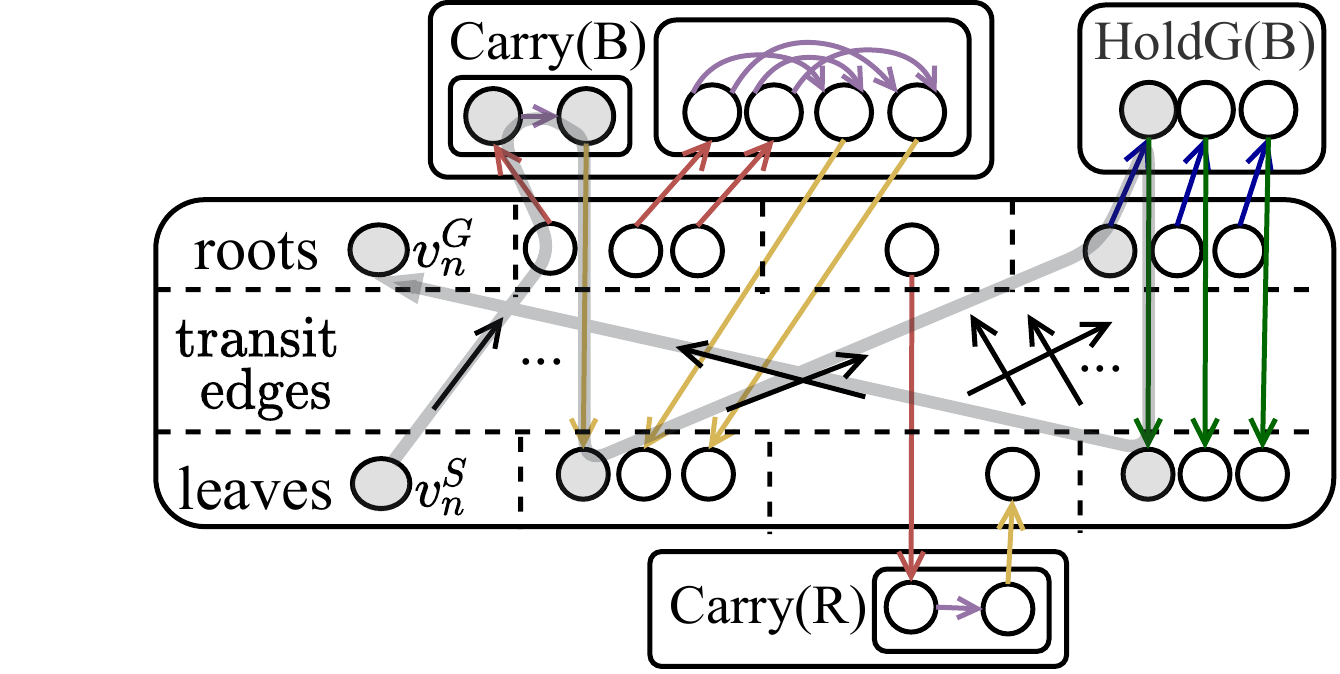}} \\
\subfloat[Subplan sequences: nodes denote modes and blocks denote primitives.]{\includegraphics[height=0.21\columnwidth]{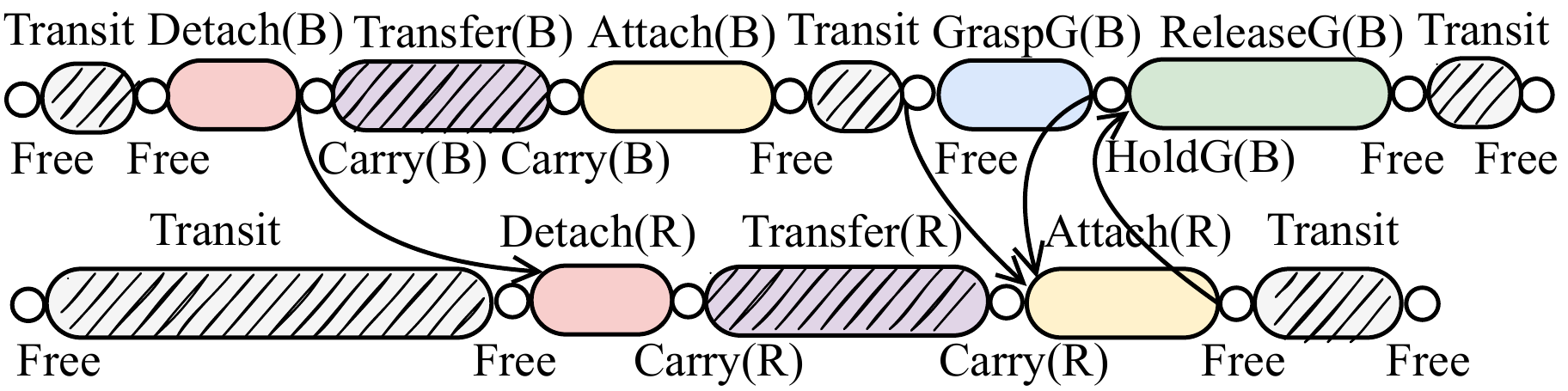}}
\caption{\small Task roadmap, plan roadmap, and subplan sequences.}
\label{fig:task_assignment}
\vspace{-16pt}
\end{figure}

\subsubsection{Task Roadmap} For every task $\textsc{T}_p = (\tau_{p,1},..,\tau_{p,K_p}) \in \textsc{T}$ and every robot $a_n$, we extract a \textit{task roadmap} $\textsc{G}^\textsc{T}_{n,p}$, an acyclic directed graph that represents all the paths to complete task $\textsc{T}_p$. We construct $\textsc{G}^\textsc{T}_{n,p}$ by (1) adding all the milestones and mode-changing edges introduced by primitives $\textsc{T}_p$ to $G_n$; and (2) in each mode, adding a mode-preserving edge from every milestone that ends a mode-changing edge to every milestone that starts a mode-changing edge with the minimum traversal time as the weight except for mode Free. We compute these weights by using the Floyd-Warshall algorithm. The added edge is an abstraction of the highways between them and thus has the same primitive. 
A task roadmap example is in Fig.\ref{fig:task_assignment}(a). 

\subsubsection{Plan Roadmap} For every robot $a_n$, we construct a \textit{plan roadmap} $\textsc{G}^\textsc{T}_n$  to represent all the paths of $a_n$ to complete tasks $\textsc{T}$. To construct $\textsc{G}^\textsc{T}_n$, we first compose all the task roadmaps $\otimes_p \textsc{G}^\textsc{T}_{n,p}$. We find the root vertices of all the task roadmaps along with the robot goal vertex as the plan roots, and the leaf vertices along with the robot start vertex as the plan leaves. Roots and leaves have zero in-degree and out-degree respectively. Then, we add an edge from every leaf to every root with its traversal time except for the vertices belonging to the same task roadmap. The nodes of this plan roadmap are not necessarily a subset of the multi-modal roadmap since some tasks are required to executed more than once. An example of plan roadmap is given in Fig.\ref{fig:task_assignment}(b).

\subsubsection{MILP Encoding} For every robot $a_n \in A$ and every edge $e \in \textsc{G}^\textsc{T}_n.\textsc{E}$, we create a binary variable $A[e]$ to indicate that $a_n$ traverses edge $e$ and a non-negative real variable $t_n[v]$ to denote the times of $a_n$ arriving $v \in \textsc{G}^\textsc{T}_n.\textsc{V}$. For every primitive $\tau_{p,k} \in \textsc{T}_p$, we use real variables $t_\textsc{T}[\tau_{p,k},\vdash]$ and $t_\textsc{T}[\tau_{p,k},\dashv]$ to denote the start and end times of $\tau_{p,k}$. 
Let $t_\textsc{T}[e.\texttt{s}]$ and $t_\textsc{T}[e.\texttt{e}]$ also denote the same variables as  $t_\textsc{T}[\tau_{p,k},\vdash]$ and $t_\textsc{T}[\tau_{p,k},\dashv]$ for $e \in \textsc{G}^\textsc{T}_n.\textsc{E}$ if $e$ is labeled with $\tau_{p,k}$.
We define a real variable $t$ to denote the total makespan. For a vertex $v \in \textsc{G}^\textsc{T}_n$, we denote its incoming edges in $\textsc{G}^\textsc{T}_n$ as $\textsc{in}(v)$ and the outgoing edges as $\textsc{out}(v)$. The implication logical operator in the MILP model can be compiled to linear constraints by using the big-M method \cite{griva2009linear}.


\begin{figure}[ht]
\vspace{-15pt}
{\fontsize{8.5pt}{12pt}
\begin{flalign}
\setstretch{1.5}
    \nonumber& \textsc{minimize} \,\,\, t & 
\end{flalign}}
\vspace{-25pt}
\end{figure}

\begin{figure}[ht]
\vspace{-20pt}
{\fontsize{8.5pt}{12pt}
\begin{flalign}
\setstretch{1.5}
    &  \textstyle{\sum}_{e \in \textsc{in}(v)}A[e] = \textstyle{\sum}_{e \in \textsc{out}(v)}A[e], \ \;\;\;\;\;\;\;\;\forall v \in \cup_n \textsc{G}^\textsc{T}_n.\textsc{V} / \{v^\textsc{S}_n, v^\textsc{G}_n\} \\
    &  \textstyle{\sum}_{e \in \textsc{out}(v)}A[v^\textsc{S}_n] = 1, \textstyle{\sum}_{e \in \textsc{in}(v)}A[v^\textsc{G}_n] = 1, \ \;\;\;\;\;\;\;\;\;\;\;\;\forall n \in (1..N) \\
    & A[e] {\implies} (t_n[e.\texttt{e}] - t_n[e.\texttt{s}] \geq e.\texttt{w}), \ \;\;\;\;\;\;\;\;\;\;\;\;\;\;\;\;\;\;\;\;\forall  e \in \cup_n \textsc{G}^\textsc{T}_n.\textsc{E} \\
    & \textstyle{\sum}_{e \in \textsc{E}} A[e] = 1, \ \:\;\;\;\;\;\;\;\;\;\;\;\;\;\;\;\;\;\;\forall \textsc{T}_{k,p} \in \textsc{T} \text{ where } \textsc{E} = \cup_n G^\textsc{T}_{n,p,k}.\textsc{E} \\
    & A[e] {\implies} (t_\textsc{T}[e.\texttt{s}] = t[e.\texttt{s}])) \land (t_\textsc{T}[e.\texttt{e}] = t[e.\texttt{e}])), \ \;\;\;\;\forall e \in \textsc{G}^\textsc{T}_{n}.\textsc{E} \\
    &t_\textsc{T}[a] \leq t_\textsc{T}[b], \ \;\;\;\;\;\;\;\;\;\;\;\;\;\;\;\;\;\;\;\forall (a, b) \in \textsc{P} \text{ and } (t \geq t_n[\cdot] \land t \geq t_\textsc{T}[\cdot]) \\
    \nonumber&  (A[e] \land A[e']) {\implies}  (t_n[e.\texttt{e}]<t_{n'}[e'.\texttt{s}]) \lor (t_{n'}[e'.\texttt{e}]<t_{n}[e.\texttt{s}])), \\
    & \ \;\;\;\;\;\;\;\;\;\;\;\;\;\;\;\;\;\;\;\;\;\;\;\;\;\;\;\;\forall (a_n, e), (a_{n'}, e') \in \Pi \text{ and } e \in G_n, e' \in G_{n'} \\
    & A[e] {\implies} (t_{n}[e.\texttt{s}]>t_\textsc{T}[\tau, \vdash]), \ \;\;\forall (a_n, e), (o, \bot) \in \Pi, \forall \tau \text{ detach } o \\
    & A[e] {\implies} (t_{n}[e.\texttt{e}]<t_\textsc{T}[\tau, \dashv]), \ \;\;\forall (a_n, e), (o, \top) \in \Pi, \forall \tau \text{ attach } o
\end{flalign}}
\vspace{-20pt}
\end{figure}

Constraints (1-2) ensure every robot $a_n$ traverses through a valid path in its plan roadmap $G^\textsc{T}_n$ from start $v_n^\textsc{S}$ to goal $v_n^\textsc{G}$, and (3) enforces the arrival times of the vertices on this path to respect traversal times. (4) constrains each task to be assigned to exactly one robot, and (5) links the start and end times of each task primitive with the arrival times of its assigned robot. (6) enforces these start and end times to satisfy the precedence constraints $\textsc{P}$, and makespan $t$ an upper bound on all the time variables. Thus, constraints (1-6) ensure the robot paths complete all the tasks on time while respecting the roadmap connectivity and traversal times.

Constraints (7-9) prevent robots from colliding with objects or each other when taking mode-changing edges,  which are often manipulation actions and more likely lead to dead ends. Constraint (7) enforces the mode-changing edges that collide with each other not to happen concurrently, and (8) guarantees such edges that collide with objects at starts or goals are not taken before detaching or after attaching objects respectively.
These constraints are paramount to producing task assignments that admit full collision-free motion plans and empirically mitigating the incompleteness of our hierarchical design. For example, in \Cref{fig:diagram}, the robot on the right can only grasp object R from the top. 
If the MILP did not have constraint (9), the planner might assign this robot to assemble object R. However, since B, which is attached before R, blocks R from being top grasped, this task assignment admits no feasible motion plan.
Indeed, in practice, without these constraints, the system fails to produce a solution for most assembly problems of interest.

\subsubsection{Extracting Subplan Sequences} A MILP solution produces a path for each robot, see the gray shadow line in  Fig.\ref{fig:task_assignment}(b). 
From a solution, we extract a \textit{subplan sequence} $\gamma_n = \{g_{n,k}\}_k$ for each robot as in Fig.~\ref{fig:task_assignment}(c). Each subplan is a mode or primitive associated with the start and end vertices as assigned in the MILP solution. A subplan $g_{n,k}$ is \textit{planned} if a time-stamped path from its start vertex to its end vertex on roadmap $G_n$ is provided. The subplans can be classified as three types: (1) a mode subplan has the same start and end vertices and thus its path is a single vertex; (2) implicitly, the path of a mode-changing primitive subplan can only traverse its corresponding edge in the MILP solution; and (3) the other primitive subplans need to plan longer paths such as Transit and Transfer primitives as sketched in Fig.~\ref{fig:task_assignment}(c). We extract such subplan sequences for all robots to obtain $\Gamma = \{\gamma_n\}_n$ and add precedence constraints $\mathcal{P}$ between the subplan end times according to the original assembly requirement. 
Then, these assigned, unscheduled, partially ordered subplans become the goal description of our path finding problem in Section~\ref{section:path}.

\begin{algorithm}[t!]
\caption{\small PBS-AT}\label{alg:pbs}
\setstretch{0.9}
\fontsize{8.5}{12}
    $\emph{Root} = (\emph{paths}, C, \PREC) \gets (\emptyset, \emptyset, \mathcal{P})$\label{line:pbs:root}\;
    $S \gets \{\emph{Root}\}$\;\label{line:pbs:init}
    \While{$(N \gets S.\emph{pop}()) \neq \emptyset$ \label{line:pbs:pop}}
    {
        \While{$\mathcal{C}_N = \emptyset$\label{line:pbs:empty-collision}}
        {
            \If{$g_i \gets \emph{NextUnplannedSubplan}( \PREC_N$) is $\False$\label{line:pbs:nextplan}}
            {\Return $N.\emph{paths}$\label{line:pbs:goal}\;}
            \If{$\emph{paths} \gets \emph{PlanPaths}(N, g_i)$ is $\False$}
            {\textbf{go to} Line \ref{line:pbs:pop}\label{line:pbs:prune}\;}
            \lForEach{$p_i \in \emph{paths}$}
            {$N.\emph{paths}[g_i] \gets p_i$\label{line:pbs:update}}
            $\mathcal{C}_N \gets \emph{UpdateCollisions}(N, paths)$\;\label{line:pbs:update-collision}
        }
        $g_i, g_j \gets$ subplans involved in the 1st collision in  $\mathcal{C}_N$\;\label{line:pbs:choose_conflicts}
        \ForEach{$(g_k, g_l) \in \{(g_i, g_j), (g_j, g_i)\}$\label{line:pbs:generate_start}}
        {
            $N' \gets (N.\emph{paths}, \mathcal{C}_N, \PREC_N \cup \{g_k \prec g_l\})$ \label{line:pbs:new_child}\;
                \lIf{$\textrm{UpdateNode}(N', g_k)$ is \True} {%
                    $S$.\textit{insert}($N'$)%
                }\label{line:pbs:generate_end}
        }
    }
    \Return $\False$\;
\end{algorithm}

\section{Path Finding with Assigned Tasks}
\label{section:path}
We now introduce our Priority-Based Search algorithm for multi-robot path finding with Assigned Tasks (PBS-AT), which plans paths for every robot $a_n$ to fulfill its subplan sequence $\gamma_n$ on its multi-modal roadmap $G_n$ such that the precedence constraints $\mathcal{P}$ are satisfied and the paths are collision-free given annotated collision $\Pi$. The idea of PBS-AT is to divide a multi-robot problem into single-robot sub-problems and explore the priorities of planning sub-problems as proposed in PBS with Precedence Constraints (PBS-PC) \cite{zhang2022multi}. PBS-PC is a two-level algorithm that plans collision-free paths for multiple robots from their starts to visit sequences of precedence-constrained goals in grid graphs ({\it i.e.,} MAPF-PC). As its high level explores the priorities between goals in a Priority Tree (PT) such that the robots that move towards lower-priority goals should avoid colliding with those that move towards higher-priority goals, its low level uses A* to plan single-robot paths optimally in discretized timesteps by reserving the paths of higher-priority goals as moving obstacles. 
Similar to PBS-PC, PBS-AT is also a two-level algorithm that explores the priorities between goals (subplans) and calls its low level to plan paths for subplans. Additionally, its low-level plans paths for multiple successive same-robot subplans at once instead of one subplan at a time. 
For simplicity, in this section, we use $g_i$, $i = 1, \ldots, \sum_{a_n \in A} |\gamma_n|$, instead of $g_{n,k}$ to denote a subplan, where subplans for different robots have different index values $i$, and $p_i$ to denote its corresponding path.

The high level of PBS-AT (Algorithm~\ref{alg:pbs}) performs a depth-first search on the PT.
It starts with the root PT node that contains an empty set of paths, an empty set of collisions, and the initial priority orderings $\PREC$, which are initialized with respect to the precedence constraints $\mathcal{P}$ to enforce subplans that end later to have lower priorities (Line~\ref{line:pbs:root}). The precedence constraints between subsequent same-robot subplans are trivially included. Then, a stack $S$ is initialized with the root node (Line~\ref{line:pbs:init}). When expanding PT node $N$ (Line~\ref{line:pbs:pop}), it first plans paths for unplanned subplans one at a time with respect to the priority orderings $\PREC_N$ (\Crefrange{line:pbs:empty-collision}{line:pbs:update-collision}), {\it i.e.} \emph{NextUnplannedSubplan} always returns the unplanned subplan that does not have any unplanned higher-priority subplans (\Cref{line:pbs:nextplan}), until (1) some collisions are found (\Cref{line:pbs:empty-collision}), (2) all paths are planned, in which case we return the paths (\Cref{line:pbs:goal}), or (3) no paths exist, in which case we prune $N$ (Line~\ref{line:pbs:prune}). 
\emph{PlanPaths}($N, g_i$) returns a set of paths because it plans a path for $g_i$ and, if necessary, replans paths for the previous same-robot subplans of $g_i$. 
Last, PBS-AT resolves a collision in $\mathcal{C}_N$ in the same way as PBS-PC and replans the paths in each generated child node by calling \emph{UpdateNode} (\Crefrange{line:pbs:choose_conflicts}{line:pbs:generate_end}). 

\setlength{\textfloatsep}{0pt}
\begin{algorithm}[t!]
\caption{\small{UpdateNode (Node $N$, subplan $g_i$)}}\label{alg:updateNode}
\setstretch{0.9}
\fontsize{8.5}{12}
    $\emph{R} \gets \{g_i\}$\tcp*{sort subplans to replan in in order of $\PREC_N$}\label{line:updateNode:queue}
    \While{$(g_j \gets \textit{R}.pop()) \neq \emptyset$\label{line:updateNode:while}} {
        \lIf{\textit{livelock} occurs}{\Return \ \False} \label{line:updateNode:livelock}
        \lIf{$\emph{paths} \gets \textrm{PlanPaths}(N, g_j)$ is \False}{%
            \Return \ \False
        }\label{line:updateNode:replan}
        \lForEach{$p_k \in \emph{paths}$}
        {$N.\emph{paths}[g_k] \gets p_k$\label{line:updatePlan:update}}
        $\textit{R} \gets \textit{R} \cup 
            \{g_l \mid (g_k \prec_N g_l) \land (g_k, g_l) \in \mathcal{C}_N, p_k \in \emph{paths}\} 
            \cup \{g_l \mid (g_k \text{ precede } g_l) \land  (N.\emph{paths}[g_l].\emph{T} < p_k.\emph{T}), p_k \in \emph{paths}\}$\;\label{line:updatePlan}
        $\textit{R} \gets \textit{R} \setminus \{g_k \mid p_k \in paths\}$\;\label{line:updatePlan:deleteR}
    }
    \Return \ \True\;\label{line:updateNode:exhaust}
\end{algorithm}
\setlength{\textfloatsep}{0pt}


\emph{UpdateNode} (Algorithm~\ref{alg:updateNode}) iteratively updates the paths of all the affected lower-priority subplans until all planned paths in $N.\emph{paths}$: (1) satisfy the precedence constraints, (2) do not collide with any objects, and (3) any two planned paths that have priorities in between are collision-free.  
It first constructs a priority queue $R$ to store all the subplans to replan, in which subplans are sorted according to $\PREC_N$ (\Cref{line:updateNode:queue}).
It then repeatedly calls \emph{PlanPaths} to replan 
until no more subplans need to be replanned (Line~\ref{line:updateNode:while}), a live lock occurs (Line~\ref{line:updateNode:livelock}), or a failure is reported by \emph{PlanPaths} (Line~\ref{line:updateNode:replan}). A live lock is a condition where updating a set of subplans triggers replanning of each other in a loop and leads to infinite replanning. In each iteration, when \emph{PlanPaths} replans paths successfully, PBS-AT updates $N.\textit{paths}$ accordingly (\Cref{line:updatePlan:update}), adds the lower-priority subplans that either violate the precedence constraints due to the updated times of the replanned subplans or collide with the updated paths to $R$ (\Cref{line:updatePlan}), and deletes the subplans that have been replanned in this iteration from $R$ (\Cref{line:updatePlan:deleteR}).

\emph{PlanPaths}($N, g_i)$ plans an optimal path for $g_i$ that (1) avoids collisions with the objects and the paths of higher-priority subplans; and (2) ends after the end time of any subplan that must end earlier than $g_i$. In the case that the subsequent same-robot subplans of $g_i$ already have paths, \emph{PlanPaths} replans their paths accordingly to avoid disjoining the paths of two subsequent subplans over time. Moreover, if there does not exist a path for $g_i$ or its subsequent subplans, \emph{PlanPaths} backtracks and replans the previous subplan of $g_i$. This backtracking procedure is repeated until \emph{PlanPaths} successfully finds paths for all the subplans that it has to plan, in which case it returns \emph{true}, or no more previous subplan exists, in which case it returns \emph{false}. In each backtracking iteration with newly added subplan $g_j$, \emph{PlanPaths} calls recursive SIPP (rSIPP) with $g_j$, and rSIPP will plan paths for $g_j$ and all its subsequent same-robot subplans that already have paths.
\begin{algorithm}[t]
\caption{{\small
rSIPP(Node $N$, subplan $g_i$, time $t_0$, RT $rt_i$)}}\label{alg:rsipp}
\setstretch{0.9}
\fontsize{8.5}{12}
    $T_\text{min} \gets \text{max}\{N.paths[g_j].T \mid g_j \in \Gamma$ should precede $g_i\}$\;\label{line:rsipp:time}
    generate root node at $g_i.\textit{start}$ at time $t_0$ and insert it to $Q$\;\label{line:rsipp:root}
    \While{$(n \gets Q.\textit{pop}()) \neq \emptyset$\label{line:rsipp:pop}}
    {
        \If{$n.v = g_i.\textit{goal} \land n.I.\textit{ub} > T_\text{min} $\label{line:rsipp:goal}}
        {
            $p \gets$ extract the path from $n$\;\label{line:rsipp:extract_path}
            \If{$p.T < T_{\min}$}
            {Add a wait till time $T_{\min}$ action to $p$\;\label{line:rsipp:add_wait1}} 
            $g_j \gets$ the subsequent subplan of $g_i$\;\label{line:rsipp:next_subplan}
            \lIf{$g_j$ is none or $N.paths[g_j] = \emptyset$}{\Return $\{p\}$\label{line:rsipp:return1}}
            $rt_j \gets \text{ReservationTable}(N, g_j)$\;\label{line:rsipp:reserve}
            \ForEach{$[lb, ub) \in rt_j.\textrm{SafeIntervals}[n.v]$\label{line:rsipp:every_interval}}
            {
                \If{$([lb', ub') \gets [lb, ub) \cap [p.T, n.I.\textit{ub})) \neq \emptyset$ $\land (\textit{paths} \gets \emph{rSIPP}(N, g_j, lb', rt_j)) \neq \emptyset$}
                {
                    Add a wait till time $lb$ action to $p$\; 
                    \Return $\{p\} \cup \textit{paths}$\;\label{line:rsipp:return2}
                }
            }
        }
        expand node $n$ w.r.t.\ $rt_i$ and insert its child nodes to $Q$\;\label{line:rsipp:expand}
    }
    \Return $\emptyset$\;
\end{algorithm}

\begin{figure*}[t!]
\vspace{-15pt}
\centering
\subfloat[Lego Bridge]{\includegraphics[height=0.3\columnwidth]{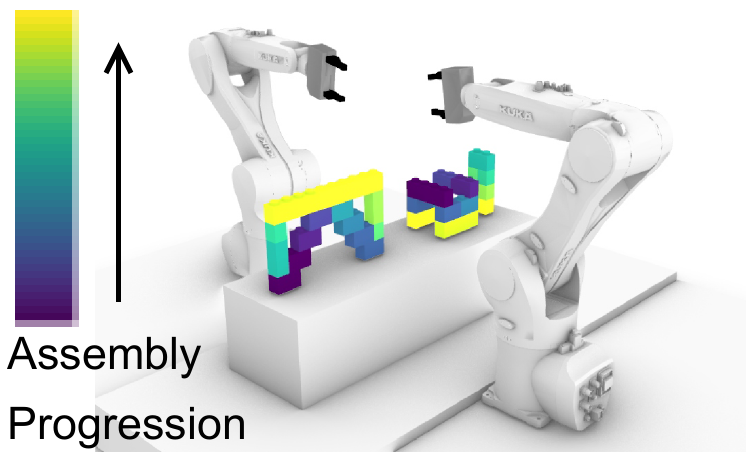}}
\,
\subfloat[Puzzle Vault]{\includegraphics[height=0.3\columnwidth]{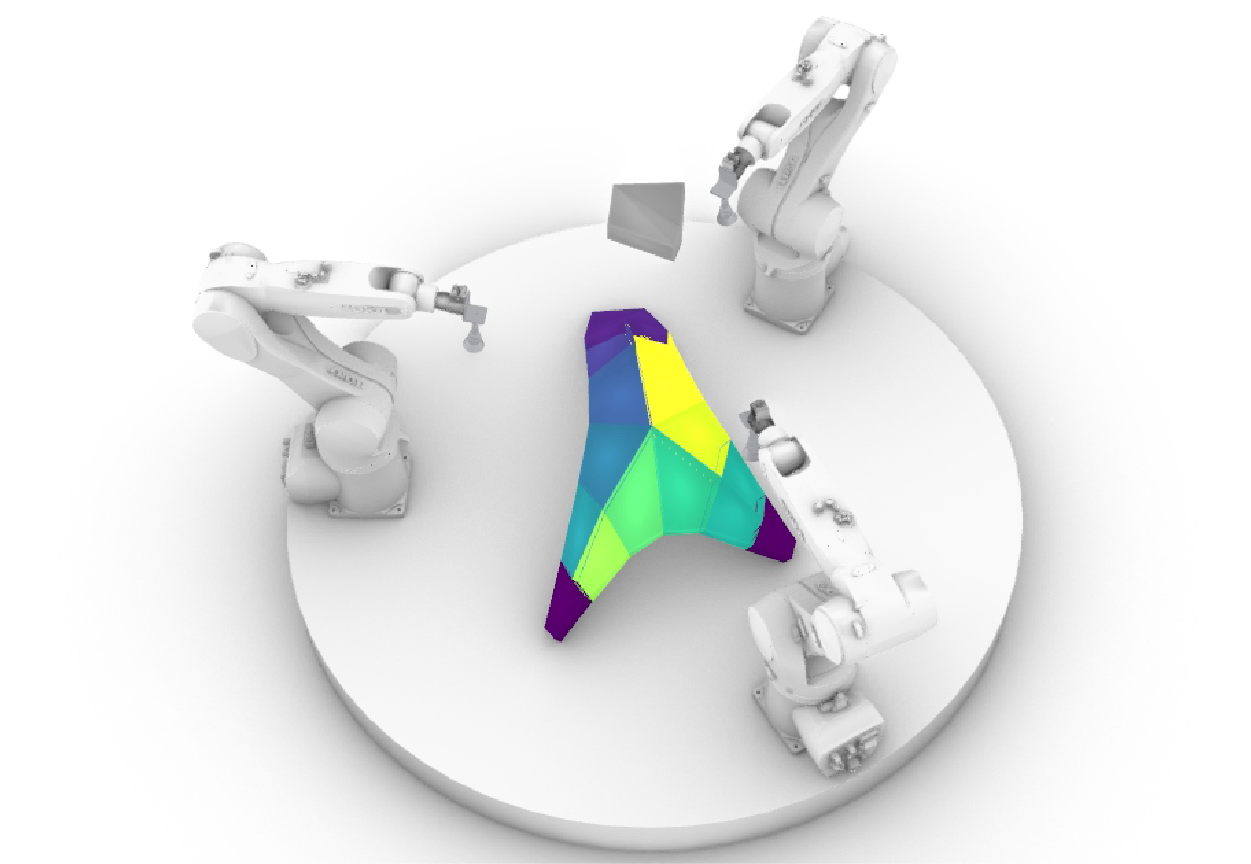}}
\,
\subfloat[Truss Boat]{\includegraphics[height=0.3\columnwidth]{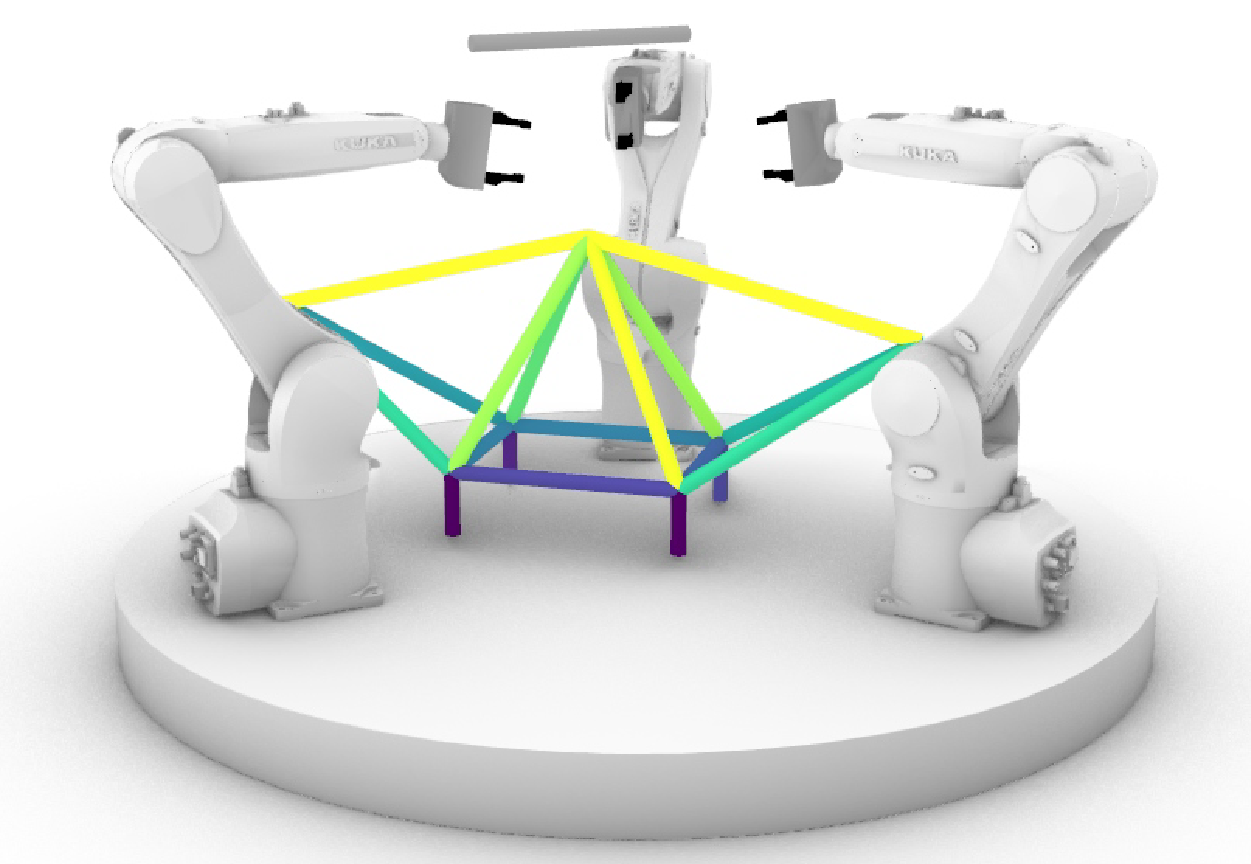}}
\,
\subfloat[Card House]{\includegraphics[height=0.3\columnwidth]{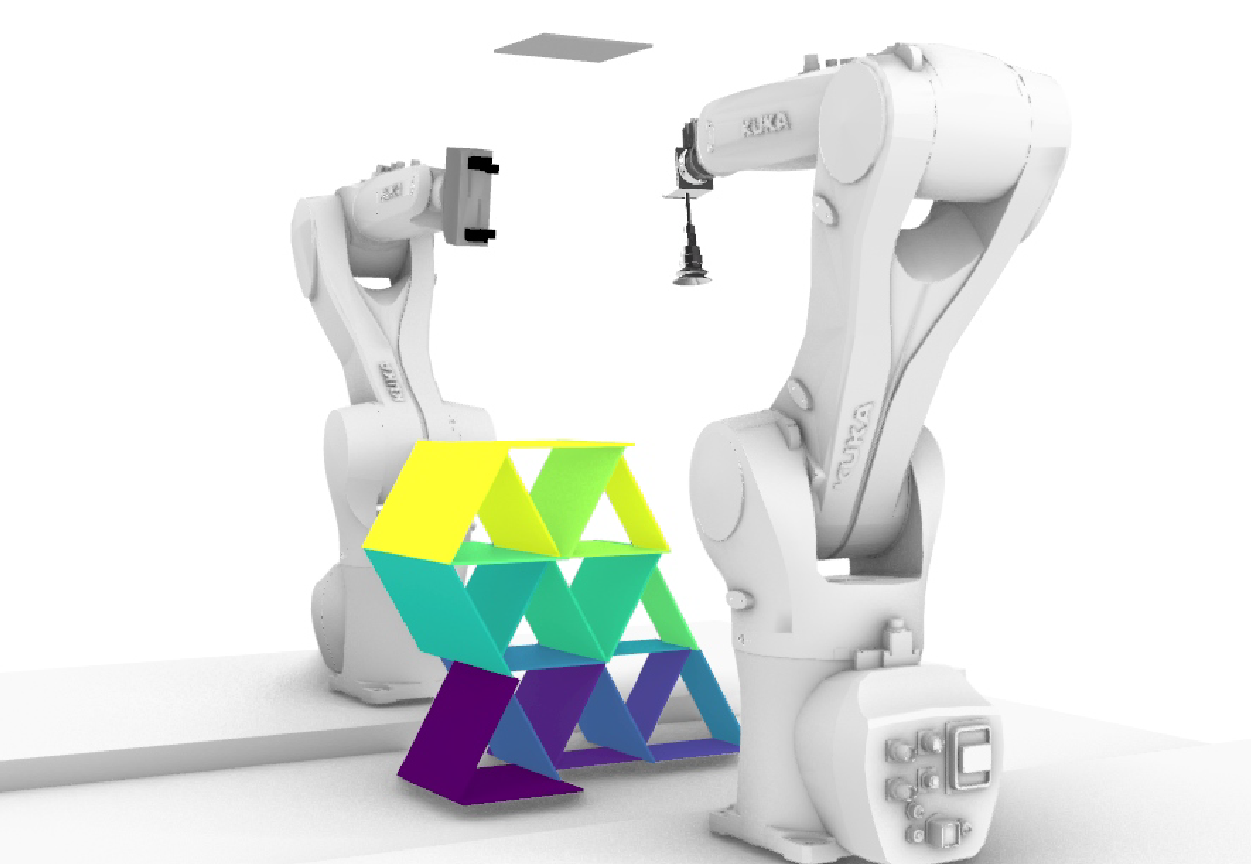}}
\vspace{-5pt}
\caption{\small Problem instances. (a) Lego Bridge: two robots with grippers assemble 17 Lego bricks; (b) Puzzle Vault: three robots with suction plates assemble 14 irregular-shaped blocks; (c) Truss Boat: three robots with grippers assemble 16 bars; (d) Card House: a robots with a gripper and a robot with a suction plate assemble 23 plates.}
\label{fig:domains}
\vspace{-1pt}
\end{figure*}

\begin{table*}[t!]\small
\centering
\begin{tabular}{|c|c|c|c|c|c|c|c|c|c|c|c|c|c|}
\hline
\multirow{2}{*}{Domain} &
  \multicolumn{4}{c|}{Roadmap Generation \& Annotation} &
  \multicolumn{4}{c|}{MILP-based Task Assignment} &
  \multicolumn{5}{c|}{PBS-AT}  \\ \cline{2-14} 
 &
  \#$V$ &
  \#$E$ &
  $t_\text{map}$ (min) &
  $t_\text{anno}$ (hr) &
  \#B &
  \#X &
  \#C &
  $t_T / t_T^\dagger / t_T^*$ (s) &
  \#$g$ &
  \#N &
  \#rSIPP &
  $t_P$ (s) &
  $\eta$    \\ \hline
(a) & 8450 & 13454 & 87 & 741  & 2985  & 83  & 95795 & 40/44/65   & 166 & 6  & 214 & 66  & 0.04  \\ \hline
(b) & 4973 & 8642  & 103 & 1289 & 1148  & 57  & 2853  & 3/10/11    & 129 & 29 & 278 & 32  & 0.22  \\ \hline
(c)   & 4049 & 8688  & 86 & 1035 & 18857 & 88  & 51224 & 90/175/496 & 153 & 44 & 221 & 122 & 0.18  \\ \hline
(d)   & 8188 & 12818 & 112 & 861  & 4253  & 141 & 13157 & 71/105/821 & 190 & 11 & 221 & 17  & 0.03 \\ \hline
\end{tabular}
\caption{\small Simulation Results. Average vertex (\#$V$) and edge (\#$E$) number of highways and connections; $t_\text{map}$: average roadmap generation time; $t_\text{anno}$: roadmap annotation time; number of binary variables (\#B), continuous variables (\#X), and constraints (\#C) in the MILP; runtimes to find a feasible MILP solution ($t_T$), find an optimal solution ($t_T^\dagger$), and exhaust the solution space ($t_T^*$);  \#g: number of subplans; \#N: number of expanded nodes; \#rSIPP: calls to rSIPP; $t_P$: PBS-AT runtime; $\eta$: ratio of the used annotated collisions.}
\label{tab:results}
\vspace{-10pt}
\end{table*}

Safe Interval Path Planning (SIPP)\cite{phillips2011sipp} is a variant of A* that finds an path \revise{with minimal total traversal time} that avoids moving obstacles. We adapt it to rSIPP (Algorithm~\ref{alg:rsipp}) so that it finds a set of paths for successive subplans that avoid the moving obstacles, {\it i.e.} the objects and the higher-priority subplans, and satisfies the precedence constraints.
It takes input a Reservation Table (RT) as in SIPP that reserves the time intervals at each vertex that are occupied by the moving obstacles.
The unreserved time intervals are called {\em safe intervals}. rSIPP searches in the resulting vertex-safe-interval graph to find an optimal path that (1) visits each vertex within a safe interval, (2) does not collide with any moving obstacles when it traverses an edge, and (3) ends no earlier than $T_{\min}$, where $T_{\min}$ is the minimum allowable time to finish this subplan with respect to all the subplans that should precede it (\Cref{line:rsipp:time}). rSIPP's has the same search procedure as SIPP except for the goal test.
When rSIPP finds a goal node (\Cref{line:rsipp:goal}), it extracts the path $p$ from $n$ (\Crefrange{line:rsipp:extract_path}{line:rsipp:add_wait1}), checks whether $g_i$ is the last subplan to replan (\Crefrange{line:rsipp:next_subplan}{line:rsipp:return1}), and if so terminates. 
Otherwise, it plans for the subsequent subplan $g_j$ to ensure a path starting from the end of $p$ exists (\Crefrange{line:rsipp:reserve}{line:rsipp:return2}). Specifically, it checks each reachable safe interval at the end vertex with respect to the RT for $g_j$, calls rSIPP for each of them, and, if succeed, returns the found paths together with $p$.

Like the PBS-PC algorithm, PBS-AT is incomplete but quite effective in practice.
Additionally, PBS-AT can explore all possible priority orderings and is biased towards solutions with a shorter makespan. 
Meanwhile, instead of only planning each subplan optimally, rSIPP also tries to optimize the paths back and forth in a non-myopic way.



\revise{\section{Simulation Results}}


In our implementation of roadmap generation and collision annotation, we leverage PyBullet Planning to generate single-mode roadmaps, check collisions, and simulate skill trajectories\footnote{\url{https://pypi.org/project/pybullet-planning/}}. We solved the MILP formulation using Gurobi \cite{gurobi2021gurobi}. The PBS-AT algorithm is implemented in C++. We tested our implementation on a 3.40GHZ 8-Core Intel Core i7-6700 CPU with 36GB RAM and leveraged 100 CPUs each with 8G memory on Amazon Web Service (AWS) to compute annotated collisions in parallel. In all the domains, we use Kuka KR-6-R900 arms with grippers or suction plates.

\revise{The skill samplers search 24 different grasps for Lego bricks and 8 for other objects to generate collision-free manipulation trajectories.} We generate 500 vertices for the Free-and-Transit roadmap by using the k-nearest neighbors PRM with $k = 10$. This generation takes roughly 40 minutes in all the domains. The maximum number of RRT-Connect samples is set to 3000. The highway vertices are connected to its 20 nearest spanned vertices via connection edges. The maximum edge duration is 0.1s, and longer edges are interpolated to sequences of edges. The joint resolution for collision checking is 0.01$\pi$ radians. 

We test our examples on domains with different features as shown in Fig.~\ref{fig:domains}: (a) Lego Bridge: two robots with grippers need to detach 17 Lego bricks from the right and assemble them as a bridge, during which necessary holding actions are provided to keep Lego bricks stable; (2) Puzzle Vault: three robots with suction plates assemble vault with 14 irregular-shaped blocks;
(3) Truss Boat: three robots with grippers assemble a bridge with 16 bars in a narrow space; (4) Card House: a {\em heterogeneous} team of two robots, one with a gripper and the other with a suction plate, cooperate to construct an assembly with 23 plates, 
In (4), the planner must account for the differing robot abilities, such as only the robot with the suction plate can place plates at the bottom. In the last three domains, we assume (1) objects spawn at the top of the construction, \revise{which abstracts the operations of some cranes or conveyors delivering the objects}, for robots to pick; \revise{(2) objects are glued to the structure when they are assembled.} 
All the domains originate the real-world designs and the precedence-constrained tasks are extracted given the design structure. 
Videos of a solution to each domain are provided in \url{https://youtu.be/hknZwLZowds}.

The runtime results and statistics of each phase (Sections \ref{section:roadmap}, \ref{section:task}, \ref{section:path}) are reported in TABLE~\ref{tab:results}. For roadmap generation and annotation, we only report the generation time of the roadmaps that are used by PBS-AT and skip the statistics for the roadmap used by task assignment. The latter roadmap of each robot is of a much smaller size and takes less than 20 minutes to generate and annotate. We report the average number of sampled vertices and edges for highways and connection edges over all the robots (\#V, \#E). 
We also report the runtime to generate this multi-modal roadmap ($t_\text{map}$) and the runtime to annotate all the necessary collision pairs ($t_\text{anno}$). This also includes the time to generate and annotate the shared configurations in Transit-Free roadmaps. As we can see in the table, the time to generate roadmaps is around 100 minutes for each robot and the annotation time can be around 1000 hours. Although we leverage CPU clusters for parallel computation which reduces the total runtime to a couple of hours,  this time can likely be significantly improved in future work by using (1) lazily roadmap generation and annotation methods \cite{bohlin2000path,li2019safe} or (2) voxel-based collision checking on a GPU\cite{lawlor2002voxel}. 

In the MILP-based task assignment, we report the numbers of binary variables (\#B), continuous variables (\#X), and constraints (\#C) along with the runtimes to find the first solution ($t_T$), find the the optimal solution ($t_T^\dagger$), and exhaust the solution space ($t_T^*$). As in TABLE~\ref{tab:results}, all the MILP encodings feature a relatively small number of continuous variables and can have up to 18000 binary variables. \#B is dominated by the number of mode-changing edges and the potential collision between these edges. Thus, we can see Truss Boat has the largest \#B since its mode-changing edges are very likely to collide with each other or the objects. Although these MILPs are large, Gurobi can find a feasible solution in two minutes and then an optimal one in a couple of minutes for all the domains, which is due to the relatively small number of continuous variables in all the domains. 

In the PBS-AT column, we show the number of subplans (\#g), expanded nodes (\#N), and calls to rSIPP (\#rSIPP), the runtime to find the solutions ($t_P$), and the ratio of the used annotated collisions over all the annotated collisions ($\eta$). As we can see, all the domains can be solved in two minutes with up to \revise{44 node expansions}. In crowded domains such as Puzzle Vault and Truss Boat, in which robots are likely to collide, PBS-AT needs to expand more nodes and add more priorities, and $\eta$ can be up to 22\%. However, in domains where robots are facing each other and do not collide often, $\eta$ can be low as 3\%.
As we can see, our task assignment and path finding algorithms can quickly find \revise{low-makespan} solutions in a couple of minutes; the runtime is dominated by roadmap generation and collision annotation. 

\vspace{8pt}
\section{Conclusion and Future Work}
We present a task and motion planning framework that jointly plans safe, \revise{low-makespan} plans for multiple, possibly heterogeneous robot arms to assemble complex structures. We demonstrate its effectiveness in several simulated assembly domains. Future work includes (1) incorporating advanced cooperative behaviors such as handovers; and (2) improving the roadmap generation and collision annotation by leveraging lazy and GPU-based collision checking. 

\bibliographystyle{IEEEtran}
\bibliography{bib}

\end{document}